\documentclass[runningheads]{llncs}
\usepackage{graphicx}
\usepackage{soul}
\usepackage{color}

\usepackage{subcaption}
\usepackage{amssymb}
\usepackage{amstext}
\usepackage{amsmath}
\usepackage{multicol}

\usepackage{hyperref}
\usepackage{graphicx}%
\usepackage{multirow}%
\usepackage{listings}%
\usepackage{caption}
\usepackage{float}
\DeclareMathOperator*{\argmax}{arg\,max}

\begin{document}
\title{Open-World Visual Reasoning by a Neuro-Symbolic Program of Zero-Shot Symbols\thanks{Supported by TNO ERP APPL.AI program.}}
\titlerunning{Open-World Visual Reasoning}
%

\author{G.J. Burghouts, F. Hillerström, E. Walraven, M. van Bekkum, F. Ruis, J. Sijs, J. van Mil, J. Dijk, W. Meijer}
\authorrunning{G.J. Burghouts et al.}

%
\institute{TNO, 2597 AK The Hague, The Netherlands}
\maketitle              

\begin{abstract}

We consider the problem of finding spatial configurations of multiple objects in images, e.g., a mobile inspection robot is tasked to localize abandoned tools on the floor. We define the spatial configuration of objects by first-order logic in terms of relations and attributes. A neuro-symbolic program matches the logic formulas to probabilistic object proposals for the given image, provided by language-vision models by querying them for the symbols. This work is the first to combine neuro-symbolic programming (reasoning) and language-vision models (learning) to find spatial configurations of objects in images in an open world setting. We show the effectiveness by finding abandoned tools on floors and leaking pipes. We find that most prediction errors are due to biases in the language-vision model.

\keywords{Neuro-Symbolic Programming \and Open World Robotics \and Zero-shot Models \and Language-Vision Models \and Knowledge Representation.}
\end{abstract}

\section{Introduction}

Finding spatial configurations of multiple objects in images is a very relevant capability. For instance, a mobile inspection robot is tasked to localize situations of interest on an industrial site, such as abandoned tools on the floor, because they may pose a hazard to the personnel and robot itself. Once such a spatial configuration is found, proper action can be taken, such as reporting it to the operator or removing the object using the robot. On such sites, there may be many activities; as a consequence, the objects, their position, and the environment may change constantly and this may differ per site. Therefore the robot may encounter new objects, e.g., a new type of tool; new environments, e.g., the floor is made of a different material; and new relevant situations, e.g., a leaking pipe. This is a challenge of the open world: handling previously unseen objects and configurations. We consider the problem of localizing spatial configurations of interest in an open world, with a focus on situations that require an action of some kind.

To identify situations of interest, which deviate from the normal, a common approach is to use a statistical model of the sensory data \cite{statistical_survey,statistical_survey_DL,statistical_sensor}. However, the robot's goal, its context and explicit prior knowledge are not taken into account. As a consequence, the detected anomalies are not necessarily relevant for the robot's mission. Another drawback of statistical models is that they do not generalize well to new observations in the open world. They cannot be adapted quickly, because it requires a significant amount of training samples to adjust the statistical model. 

We take a different approach by leveraging prior knowledge, in this case knowledge about spatial configurations of objects. This knowledge can be adapted quickly during operation and via generic definitions it can generalize better to new situations. An example of a spatial configuration is a tool that is left behind on the floor. A tool can be one of many types, such as a hammer, screwdriver, wrench, and many more. Likewise, floors can be composed of different materials with various appearances. Our goal is to find spatial configurations in images, based on a high-level definition of the involved object (categories) and the spatial relations between them, with the ability to define the object or its category, and without learning dedicated models for each of them. The rationale is that such an approach has a broader applicability, because it can generalize better across similar configurations and is adaptable to new configurations by formulating a new definition. 

\begin{figure}[H]
    \centering
  \includegraphics[width=\textwidth]{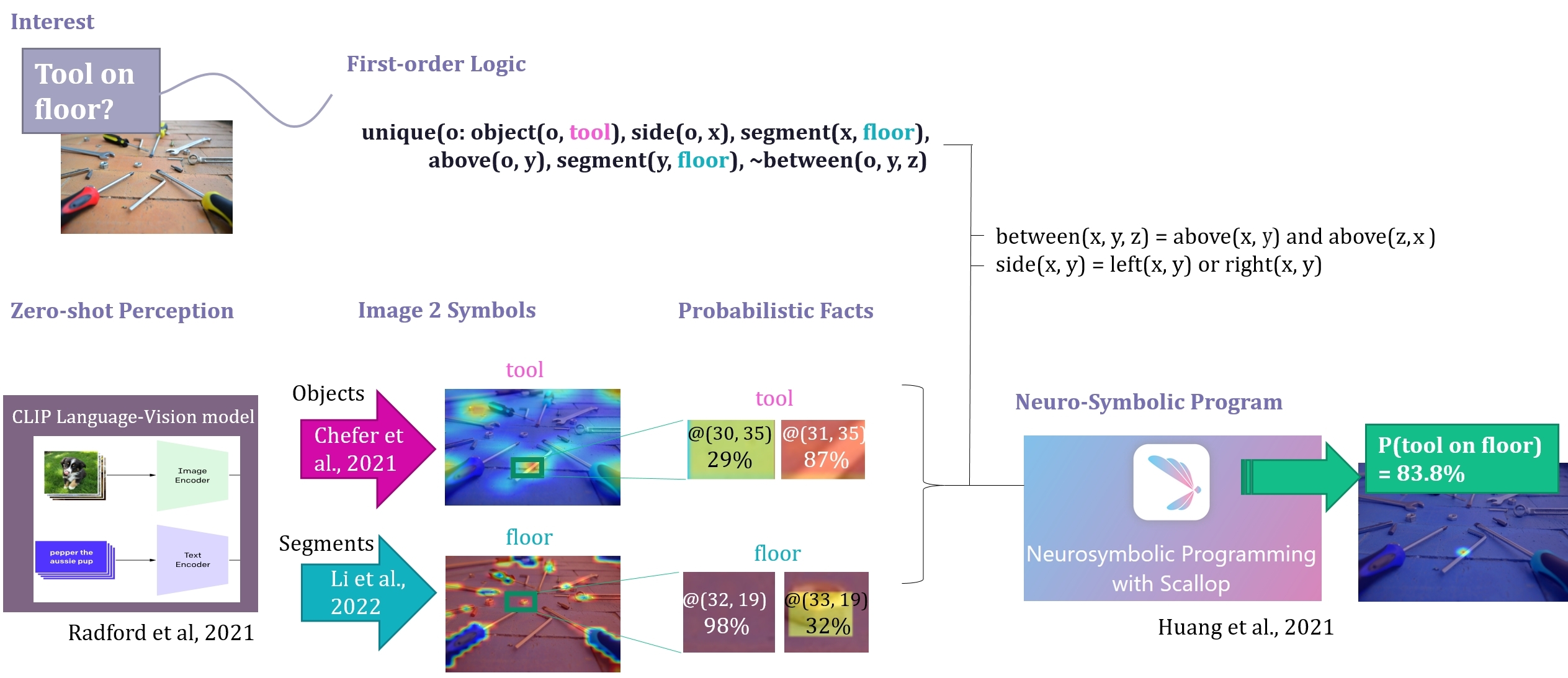}
  \caption{Finding spatial configurations of object categories in an open world. A configuration is specified by first-order logic. The symbols relate to (possibly novel) objects that are extracted from images in a zero-shot, probabilistic manner. A neuro-symbolic program validates hypotheses.}
  \label{fig:method}
\end{figure}

We start by defining the configuration of interest by first-order logic. Logic formulas specify the object categories and their spatial configuration in terms of relations between them. For flexibility and generalization, the logic formulas may entail the categories of the objects (tool), instead of the specific objects (hammer, screwdriver, etc.). This reduces the amount of human effort. The logic formulas are symbolic, where each relation and object (category) is expressed as a symbol. Each symbol is used as a query to generate object proposals for the current image. Object proposals are generated from language-vision models \cite{clip_paper,florence,align}, by querying them for the symbols that are in the logic formulas. In the image, there may be many objects and their proposals will be imperfect, leading to many possible hypotheses. Therefore, a multi-hypothesis framework is required. A natural choice is to leverage a neuro-symbolic program \cite{neuro_symbolic,hybrid_ai,scallop} for this purpose, as it validates the logic formulas against many hypotheses about the object proposals and the relations between them. Our method is outlined in Figure \ref{fig:method}. Our contribution is the integration of the neuro-symbolic programming (reasoning) with the language-vision models (learned) for the purpose of finding spatial configurations of objects. We show the effectiveness on real-world images by finding specific situations in a robotic inspection setting.

\section{Related Work}


To operate in an open world, it is essential to interpret situations well. One aspect of such interpretation is to analyze configurations of objects in the scene. A typical approach is to incorporate prior knowledge when analyzing images \cite{knowledge}. Connecting knowledge representation and reasoning mechanisms with deep learning models \cite{neuro_symbolic} shows great promise for learning from the environment and at the same time reasoning about what has been learned \cite{hybrid_ai}. Previous reasoning methods based on logic, such as DeepProbLog \cite{deepproblog,deepproblog2}, were limited in terms of scalability when there were many possible hypotheses. For instance, a task such as industrial inspection involves many possible objects and relations and therefore many possible hypotheses. Therefore, such methods were ill-suited for real-world applications. A more efficient variant of DeepProbLog was proposed \cite{deepproblog_efficient}. Recently, a framework was proposed that further improved the efficiency: the neuro-symbolic programming framework called Scallop \cite{scallop}. Scallop is based on first-order logic and introduces a tunable parameter $k$ to specify the level of reasoning granularity. It restrains the validation of hypotheses by the top-$k$ proofs. This asymptotically reduces the computational cost while providing relative accuracy guarantees. This is beneficial for our purpose, as we expect many possible hypotheses in complex environments with many objects and imperfect observations. In \cite{scallop}, the neuro-symbolic programming framework was used to reason about visual scenes using a pre-defined set of classes of objects. End-to-end learning between objects and a neuro-symbolic program was proposed to jointly learn visual concepts, words and semantic parsing \cite{mao2019neuro}. Both approaches rely on a fixed set of visual concepts, since a vocabulary, knowledge graph, or learning scheme are involved. We aim to generalize to the open world, extending the vocabulary of logical symbols to previously unseen objects, and enabling one to define symbols at the category level instead of the class level. 

For open world settings, perception models have to be applicable for a wide variety of tasks in a broad range of settings. General-purpose vision systems have been proposed, e.g., \cite{generalpurpose}, that are trained on a large set of datasets and tasks. Because some of these models can solve various tasks at the same time, these are coined foundation models \cite{foundation}. An example of such a complex task is where the model provides answers to textual questions about images \cite{knowledge,vqa}. Large progress has been achieved in language-vision tasks. Language-vision models learn directly from huge datasets of texts describing images, which offers a broad source of supervision \cite{clip_paper,clip_blog,florence,align}. They have shown great promise to generalize beyond crisp classes and towards semantically related classes. This so-called zero-shot capability is beneficial for recognizing the object categories that are involved in the spatial configuration of interest. Recently, these models were extended with capabilities to localize objects in images via co-attentions \cite{clip_attention} and to segment parts of the scene based on textual descriptions \cite{clip_segmentation}. We adopt both methods to relate image parts to respectively objects and segments from the environment, which are relevant for the configuration at hand. 

The abovementioned works in language-vision models have made huge progress in learning visual concepts in relation to language and semantics. However, they have not considered a combination with reasoning. We integrate neuro-symbolic programming with language-vision models.To the best of our knowledge, this work is the first to combine neuro-symbolic programming and language-vision models to find spatial configurations of objects in images in an open world setting.

\section{Method}

We find spatial configurations of object categories, based on prior knowledge about the involved objects and their relations. An overview of our method is shown in Figure \ref{fig:method}. At the top, it shows how a configuration of interest, such as our working example `tool on floor', is translated into symbolic predicates such as $object(o,\,tool)$, $segment(x,\,floor)$ and $above(o, x)$. At the bottom left, the figure shows how the symbols from the predicates, such as `tool' and `floor', are measured from images by language-vision models. These measurements are transformed into probabilistic facts, e.g., $P(tool | image)$ and $P(floor | image)$, which are provided to the neuro-symbolic program which validates them against the logic (bottom right). Each component is detailed in the following paragraphs.

\subsection{First-order logic} 

The configuration of interest is defined by logic formulas and predicates. The symbols in the predicates are about the objects and segments in an image. For a tool that is left on the floor, the formulas are:
\begin{equation}
\begin{split}
\exists o: & object(o, tool) \land side(o, s_1) \,\, \land\\
           & segment(s_1, floor) \land above(o, s_2) \,\, \land \\
           & segment(s_2, floor) \land \lnot between_{vert}(z, o, s_2)
\end{split}
\end{equation}

This defines that a tool on the floor is defined by seeing the tool above and aside the floor. For a robot, its perspective is oblique downward, i.e., the floor will be visible at the bottom of the tool and on the side of the tool. We also define that the tool should be on the floor, without anything ($z$) in between. Otherwise, a tool that is on a cabinet standing on the floor, would also fulfil the definition. The helper predicates are:

\begin{equation}
\begin{split}
between_{vert}(o_1, o_2, o_3) & = above(o_1, o_2) \land above(o_3, o_1)\\
side(o_1, o_2) & = left(o_1, o_2) \lor right(o_1, o_2)
\end{split}
\end{equation}

to express that some $o_2$ is vertically between $o_1$ and $o_3$. Finally, these are the predicates for positioning of one object relative to the left, right and above another object:

\begin{equation}
\begin{split}
left(o_1, o_2) & = pos_x(o_1) < pos_x(o_2)\\
right(o_1, o_2) & = pos_x(o_1) > pos_x(o_2)\\
above(o_1, o_2) & = pos_y(o_1) < pos_y(o_2)\\
\end{split}
\end{equation}

Another example of a spatial configuration is a leaking pipe. Analogous to the abandoned tool, it can be formulated by predicates that relate objects and segments:

\begin{equation}
\begin{split}
\exists o: & object(o, pipe) \land neighbor(o, s) \,\, \land\\
           & segment(s, leakage)
\end{split}
\end{equation}

An additional helper predicate is needed to express that one object is neighboring another object:

\begin{equation}
\begin{split}
neighbor(o_1, o_2) = & | pos_x(o_1) - pos_x(o_2) | \leq 1 \,\, \land\\ 
                      & | pos_y(o_1) - pos_y(o_2) | \leq 1\\
\end{split}
\end{equation}

We refer to the set of logic formulas as $L$.

\subsection{Image to Symbols} 

To find a specified configuration in the current image, the symbols from the logic formulas are used to initiate corresponding image measurements. From the image, measurements are taken that are an estimate of the symbols. We refer to this process as `image to symbols'. For each logical symbol, we produce a probabilistic fact about that symbol for the given image, e.g., $P(tool | image)$ and $P(floor | image)$. The probabilistic facts are input to the neuro-symbolic inference process (next subsection). 

Each symbol generates probabilistic proposals for the image by prompting a language-vision model. For our purpose, we adopt CLIP \cite{clip_paper} because we found its performance to be solid for various symbols, even for niche objects (i.e., zero-shot performance \cite{clip_blog}). We query CLIP by a text prompt of the symbol, e.g. `tool'. CLIP operates on the image level, i.e., matching a prompt to the full image. Our objective is different: we aim to localize the symbols, such that we can analyze their spatial configurations within the image. We adopt recent extensions of CLIP that can generate attention maps \cite{clip_attention} and segmentation maps \cite{clip_segmentation} for text prompts. The attention maps show where the model is activated regarding objects, as the model is specialized to learn the important objects in images. Hence we use them to localize objects (e.g., tool). However, this model is object-specific and not suitable for environmental elements such as walls and floors. For this purpose, we leverage a segmentation model to segment scenes, in order to localize environmental segments (e.g., floor). In this way, we obtain probabilistic symbols. 

The attention maps for the CLIP model are obtained by \cite{clip_attention}:

\begin{equation} \label{attention}
\overline{A} = \mathbb{E}_h((\nabla A \odot A)^+)
\end{equation}

where $\odot$ is the Hadamard product, $\nabla A = \frac{\partial y_k}{\partial A}$ is CLIP's output for a textual prompt $T_k$, $\mathbb{E}_h$ is the mean across CLIP's transformer heads, and $\cdot^+$ denotes removal of negative contributions \cite{clip_attention}. 

$\overline{A}$ is an attention map and therefore it is not calibrated to probabilistic values. For that purpose, we scale $\overline{A}$ to the range $[0, 1]$ by dividing by its maximum value, thereby we obtain $\overline{A}'$. The values of $\overline{A}'$ are not probabilities. That is, the values are uncalibrated for the prompt $T_k$, since the maximum value is always 1. To calibrate $\overline{A}'$ for a prompt $T_k$, we weight $\overline{A}'$ with the confidence for $T_k$, $Y_k$: $G(k) = \overline{A}' \odot Y_k$, with $Y = softmax(\{y\})$ for CLIP's outputs $\{y\}_{1:N}$ for the respective set of prompts $\{T\}_{1:N}$. We consider a set of prompts that contrast the objects of interest (e.g., tool) with irrelevant negatives (e.g., wall, floor, closet, ceiling, etc.): $T$ $=$ $\{tool, ..., ceiling\}$. Since $G(k)$ is calculated for an image $I$ at its pixels $(i,j)$, we rewrite: $G(I,i,j,k)$. To obtain a proxy for the probability for the object of interest $o$ at index $k$ from set $T$, we take: $G(I,i,j,k_o)$. For the full image, we denote the probability map shortly as $G(I, o)$.

To segment image $I$ at image location $(i,j)$, the response to a textual prompt $T_k$ from the set of prompts $\{T\}_{1:N}$ is given by \cite{clip_segmentation}:

\begin{equation} \label{segmentation}
f(I,i,j,k) = I'(i,j) \cdot T'_k, \,\, k \in \{1, ..., N\}
\end{equation}

where $I'$ is the visual encoding of image $I$, $T'$ the textual encoding of prompt $T$ and $\cdot$ is the dot product. The vector $f$ is transformed by the softmax operator to obtain values in the range of $[0, 1]$: $F(I,i,j,:) = softmax(f(I,i,j,:))$. We consider a set of prompts that contrast the segments of interest (e.g., floor) with irrelevant negatives (e.g., wall, ceiling, etc.): $T$ $=$ $\{floor, ..., ceiling\}$. To obtain the proxy of a probability for the segment of interest $s$ at index $k$ from set $T$, within image $I$ at pixel $(i,j)$, we take: $F(I,i,j,k_s)$. For the full image, we denote the probability map shortly as $F(I, s)$. 

Using $F(I, s)$ and $G(I, o)$, we obtain a probability for each symbol from the logic formulas, where for each pixel of image $I$ the probability is stored:

\begin{equation} \label{heatmaps}
\begin{split}
P(object = o \, | \, I, o) = &\,\, G(I, o), \,\,\, o \in \{tool, pipe, etc.\}\\
P(segment = s \, | \, I, s) = &\,\, F(I, s), \,\,\, s \in \{floor, leakage, etc.\}
\end{split}
\end{equation}

\subsection{Neuro-Symbolic Inference} 

Inference of the probability for the (spatial) configuration $C$ as defined by $L$, in the image $I$ is performed by the neuro-symbolic program. This program operates on probabilistic facts \cite{scallop} which we derive from the symbolic heatmaps $P(s)$ and $P(o)$ (Equation \ref{heatmaps}). We consider various spatial scales to infer $C$, because the distance between the robot and the scene may differ from time to time. As a consequence, the amount of pixels on the objects may vary. The original heatmaps $P(s)$ and $P(o)$ are finegrained. For multi-scale analysis, we add down-sampled versions of them, $P(s, \sigma)$ and $P(o, \sigma)$ at spatial scales $\sigma$ $\in$ $\{1, 2, .., 32\}$ to indicate the down-sampling factor. This enables both finegrained and coarser inference, to accommodate for varying distances from camera to the relevant objects. To achieve scale-invariant inference, we select the scale $\sigma$ that maximizes the probability $P(C)$ for the spatial configuration $C$ in the current image $I$ given the logic $L$ and its symbols $S$ and $O$:

\begin{equation} \label{multiscale}
P(C) = \argmax_{\sigma \in \{1, \, 2, \, 4, \, 8, \, 16\}} P(C \, | \, I, \, L, \, \{P(s, \sigma)\}_{s \in S}, \, \{P(o, \sigma)\}_{o \in O})
\end{equation}

Note that this can be generalized to an optimal scale $\sigma$ specifically for each segment $s$ and object $o$ in Equation \ref{multiscale}. However, this requires that all possible combinations of $s$ and $o$ at all scales $\sigma$ are analyzed by the neuro-symbolic program $P$. This will cause a computational burden. For computational efficiency, we optimize a single scale $\sigma$ for all $s$ and $o$.

\section{Experiments}

To analyze the performance of our method, we collected a wide variety of test images to find various spatial configurations of objects. The configurations are about undesired situations that require action: finding an abandoned tool on the floor inside a factory, and detecting a leaking pipe on an industrial site. We evaluate our method in a zero-shot manner: we use the models without any retraining so they have not been optimized for the tested situations. The purpose of this setup is to validate the method in an open world.

\subsection{Abandoned Tool on Floor}

We collected 31 test images about an abandoned tool on the floor. To validate how well the method generalizes to various tools, we include images with hammers, screwdrivers, wrenches, etc. For the same reason, we include various floors, with different materials, textures and colors. Moreover, the viewpoint and zoom are varied significantly. There are 9 images of tools on floors. These are the positives which should be detected by our method. To verify true negatives, we include 8 images where there is a both a tool and a floor, but the tool is not on the floor (but on a cabinet, wall, etc.). There are 5 images with only a floor (no tool) and 4 images with only a tool (no floor). To verify true negatives, there are also 5 images where there is no tool and no floor.

\begin{figure}[H]
    \centering
     \begin{subfigure}[b]{0.47\textwidth}
         \centering
         \includegraphics[width=\textwidth]{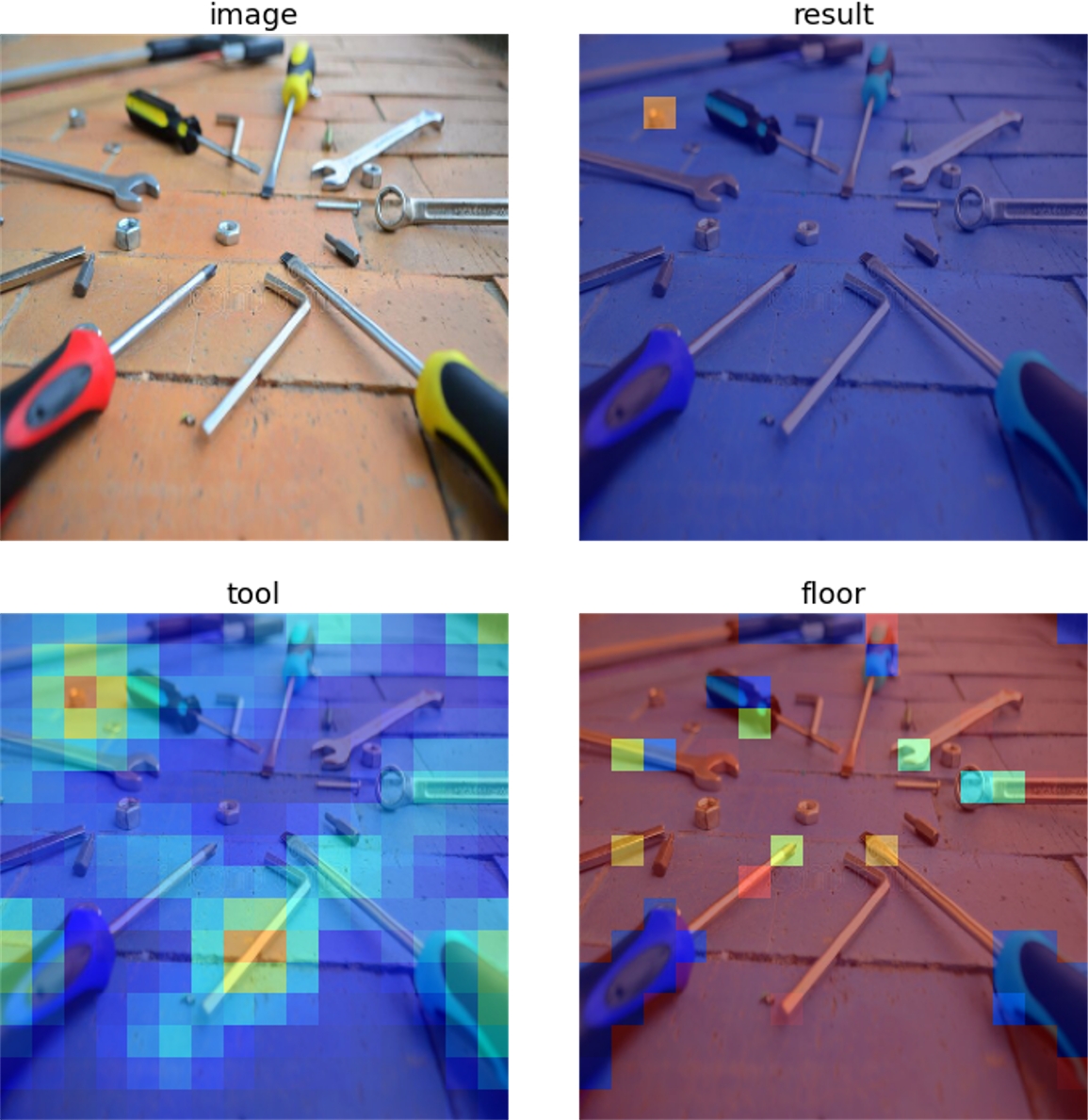}
         \caption{True positive}
         \label{fig:tool_tp_a}
     \end{subfigure}
     \hfill
     \begin{subfigure}[b]{0.47\textwidth}
         \centering
         \includegraphics[width=\textwidth]{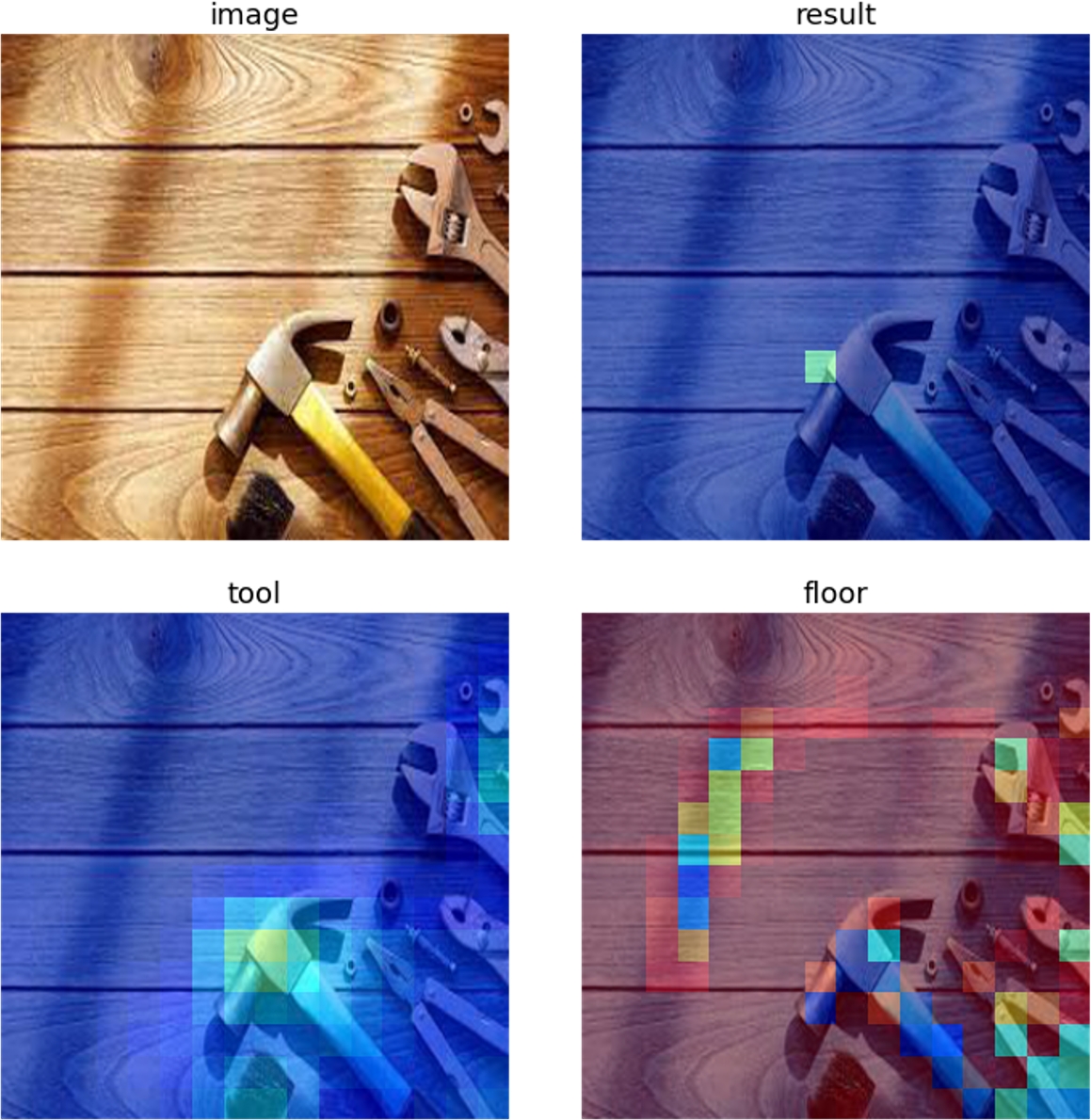}
         \caption{True positive}
         \label{fig:tool_tp_b}
     \end{subfigure}\\
     \centering
     \begin{subfigure}[b]{0.47\textwidth}
         \centering
         \includegraphics[width=\textwidth]{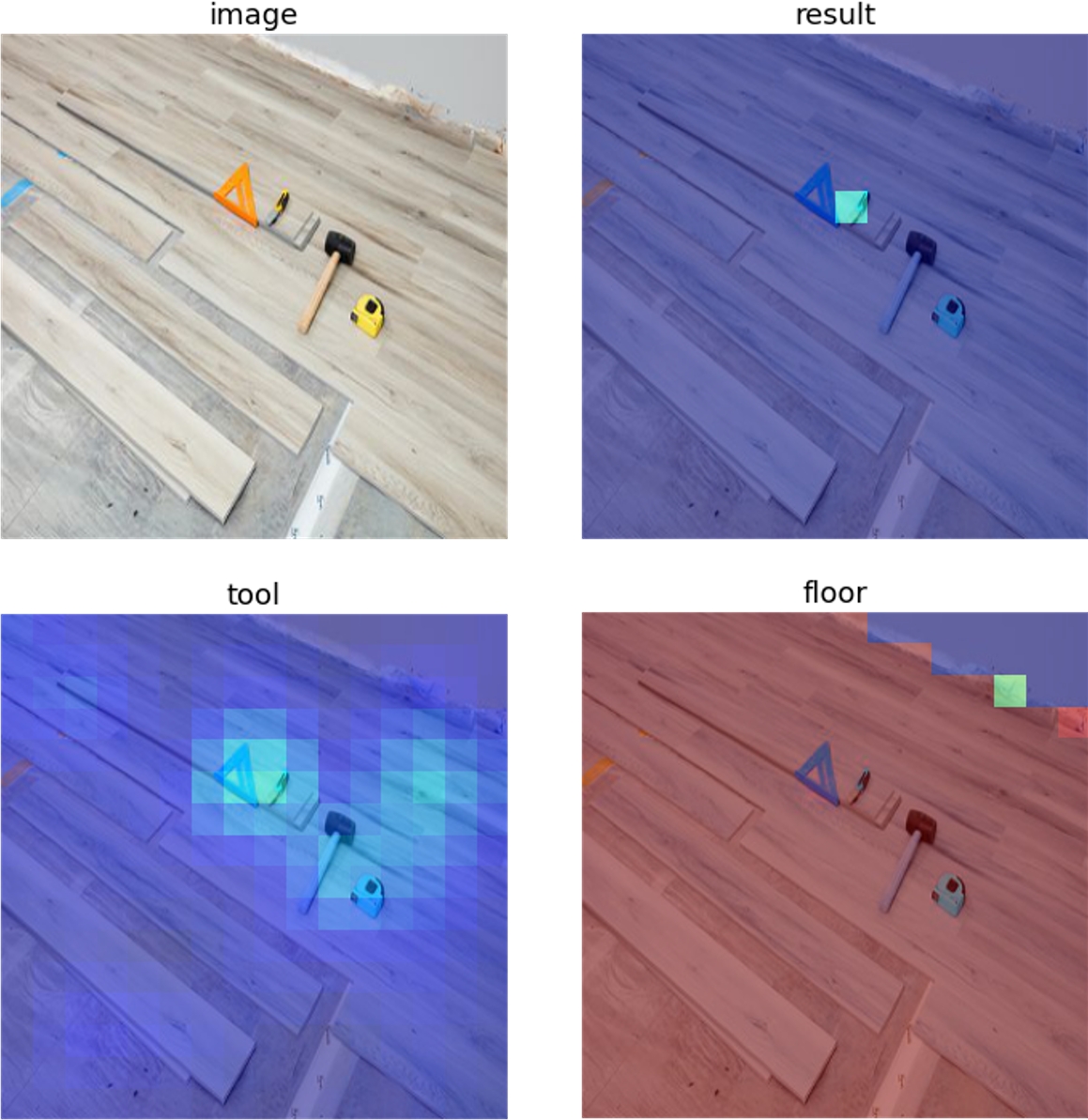}
         \caption{True positive}
         \label{fig:tool_tp_c}
     \end{subfigure}
     \hfill
     \begin{subfigure}[b]{0.47\textwidth}
         \centering
         \includegraphics[width=\textwidth]{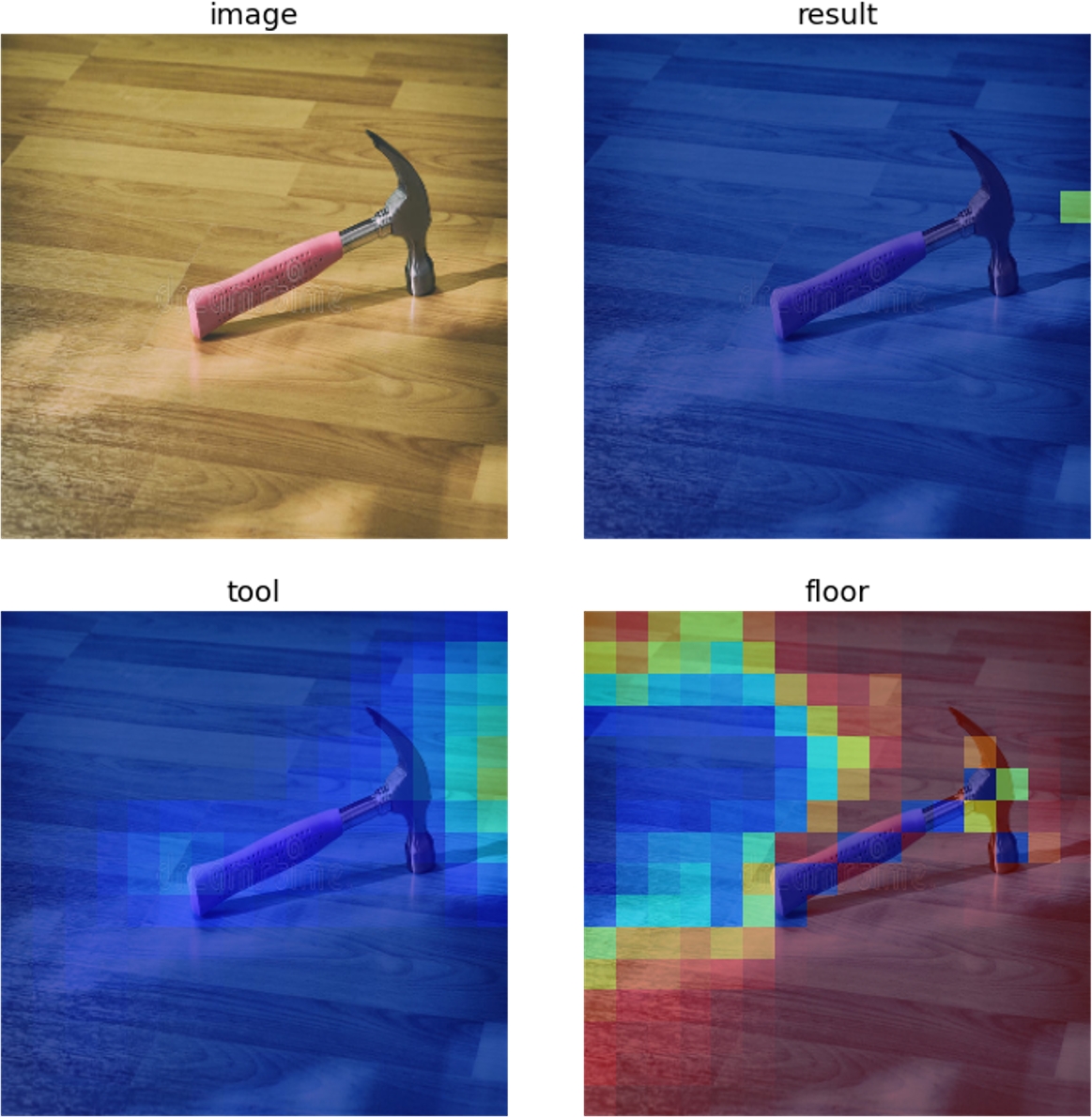}
         \caption{True positive}
         \label{fig:tool_tp_d}
     \end{subfigure}
     \\
     \centering
     \begin{subfigure}[b]{0.47\textwidth}
         \centering
         \includegraphics[width=\textwidth]{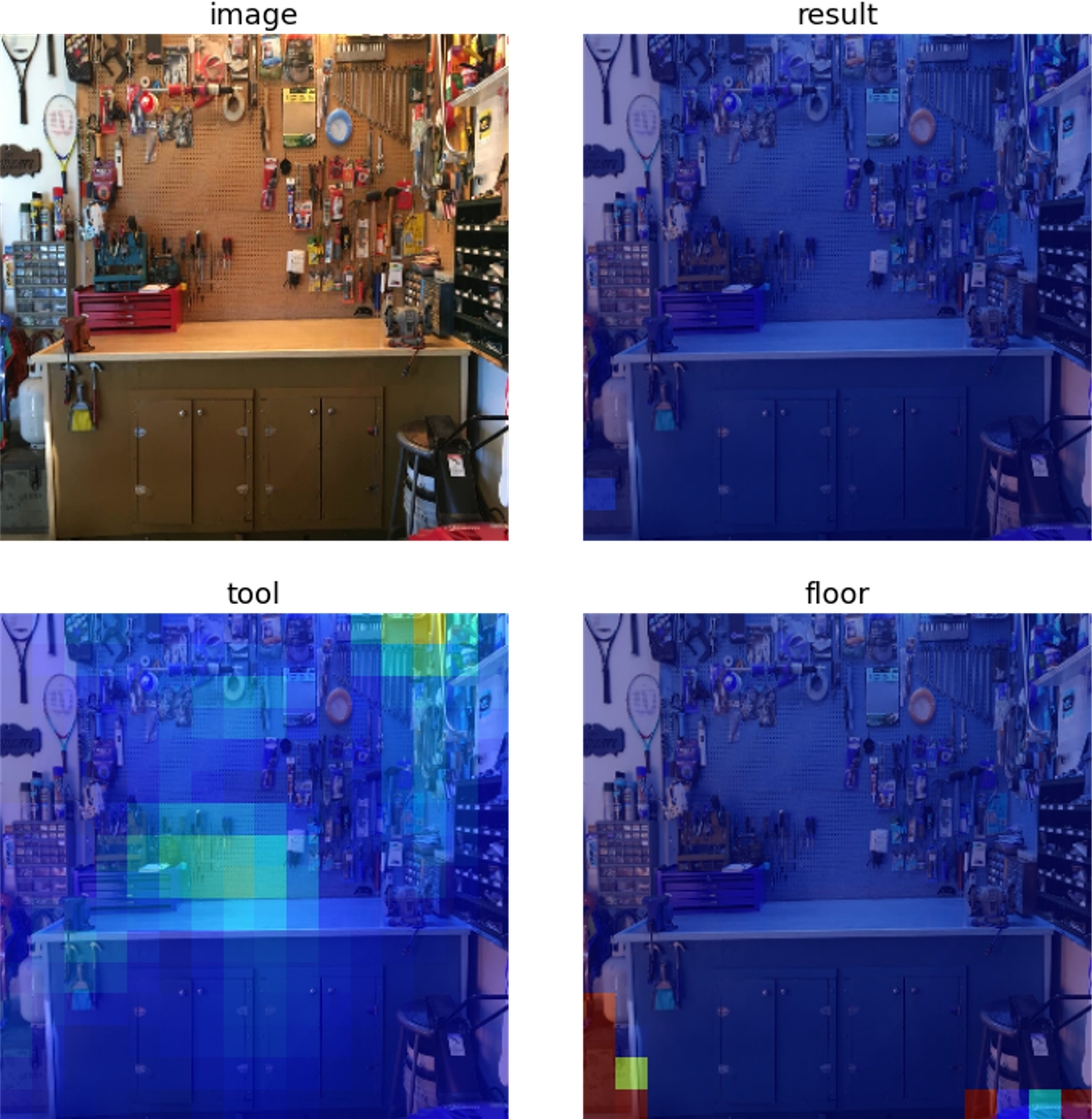}
         \caption{True negative}
         \label{fig:tool_tn_a}
     \end{subfigure}
     \hfill
     \begin{subfigure}[b]{0.47\textwidth}
         \centering
         \includegraphics[width=\textwidth]{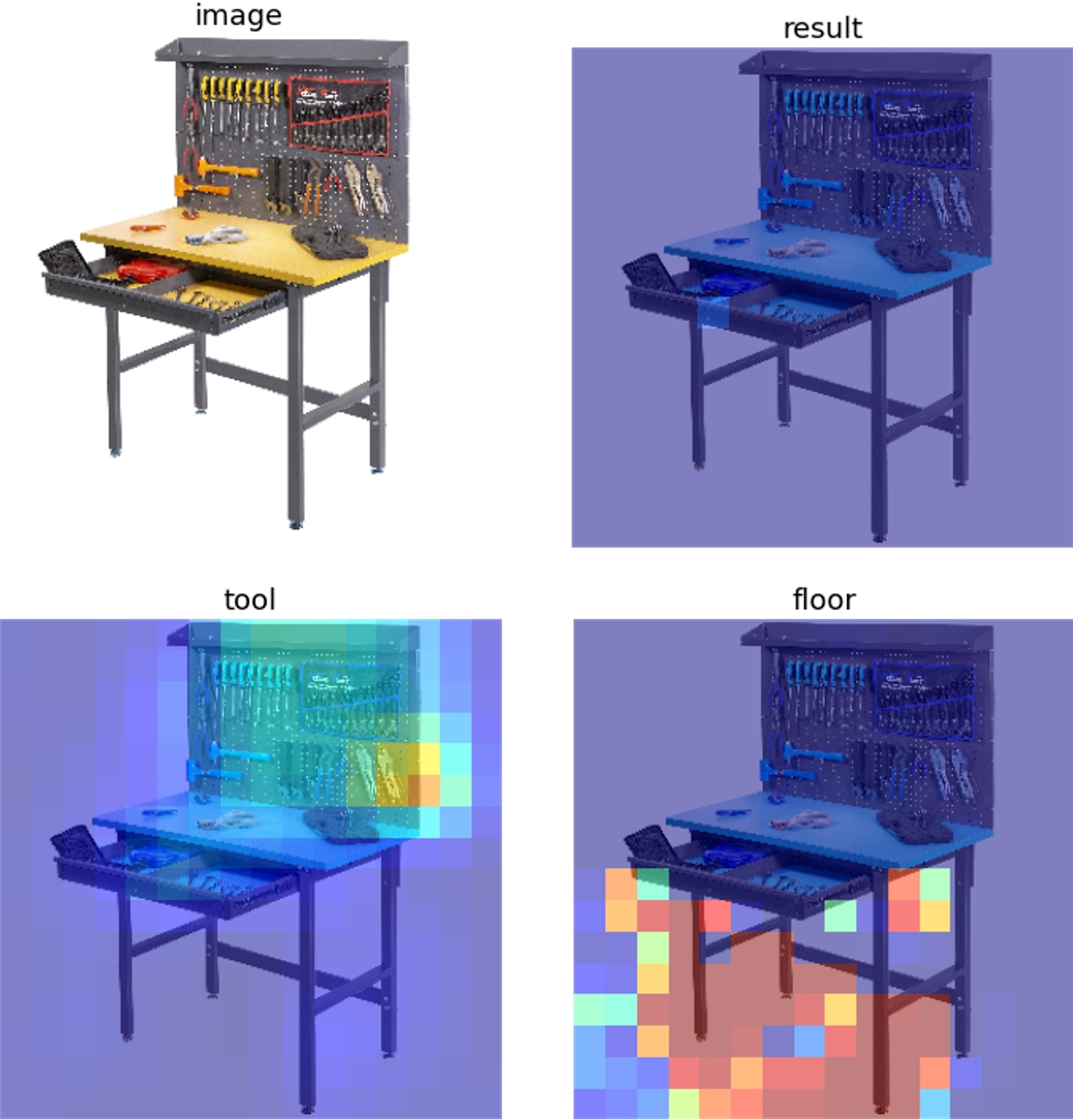}
         \caption{True negative}
         \label{fig:tool_tn_b}
     \end{subfigure}
     \caption{Tool on floor: good predictions.}
     \label{fig:tool_good}
\end{figure}

There are 9 positives, from which we detect 7 cases. Figure \ref{fig:tool_tp_a} - \ref{fig:tool_tp_d} shows 4 out of those 7 cases. Each case is illustrated by showing the original image in the top left and the result of the neuro-symbolic program in the top right. The involved symbols are shown in the lower left and right. The visualizations are heatmaps, where a high probability is red, whereas blue indicates a low probability. For the symbols (lower rows), the full heatmap is shown. For the result of the neuro-symbolic program (top-rights), only the most likely location is shown. 

The resolution of the heatmaps indicates the granularity of the reasoning, as it reflects the optimal spatial scale which maximized the resulting probability (Equation \ref{multiscale}). In Figure \ref{fig:tool_good}, most spatial scales are relatively small, indicated by the fine-grained heatmaps. Since the symbols are predicted well (often the tools and floors have a high probability at the respective symbols), the reasoning is able to pinpoint a place in the image where the spatial configuration is fulfilled mostly (a peak in the neuro-symbolic output). In the case of a negative, the probability is typically much lower, as shown in the next paragraph.

Out of the 17 negatives, there are 8 hard cases, as these images contain both a tool and a floor, but not in the defined configuration: the tool is not on the floor. Out of these 8 negatives, 6 are correctly assessed as such. The two errors (i.e., false positives) are shown in the next subsection. Figure \ref{fig:tool_tn_a} and \ref{fig:tool_tn_b} show 2 out of 6 true negatives. Although there are both a floor and tools, the reasoner correctly finds that the spatial configuration is not a tool that is on the floor.

\begin{figure}[H]
    \centering
     \begin{subfigure}[b]{0.47\textwidth}
         \centering
         \includegraphics[width=\textwidth]{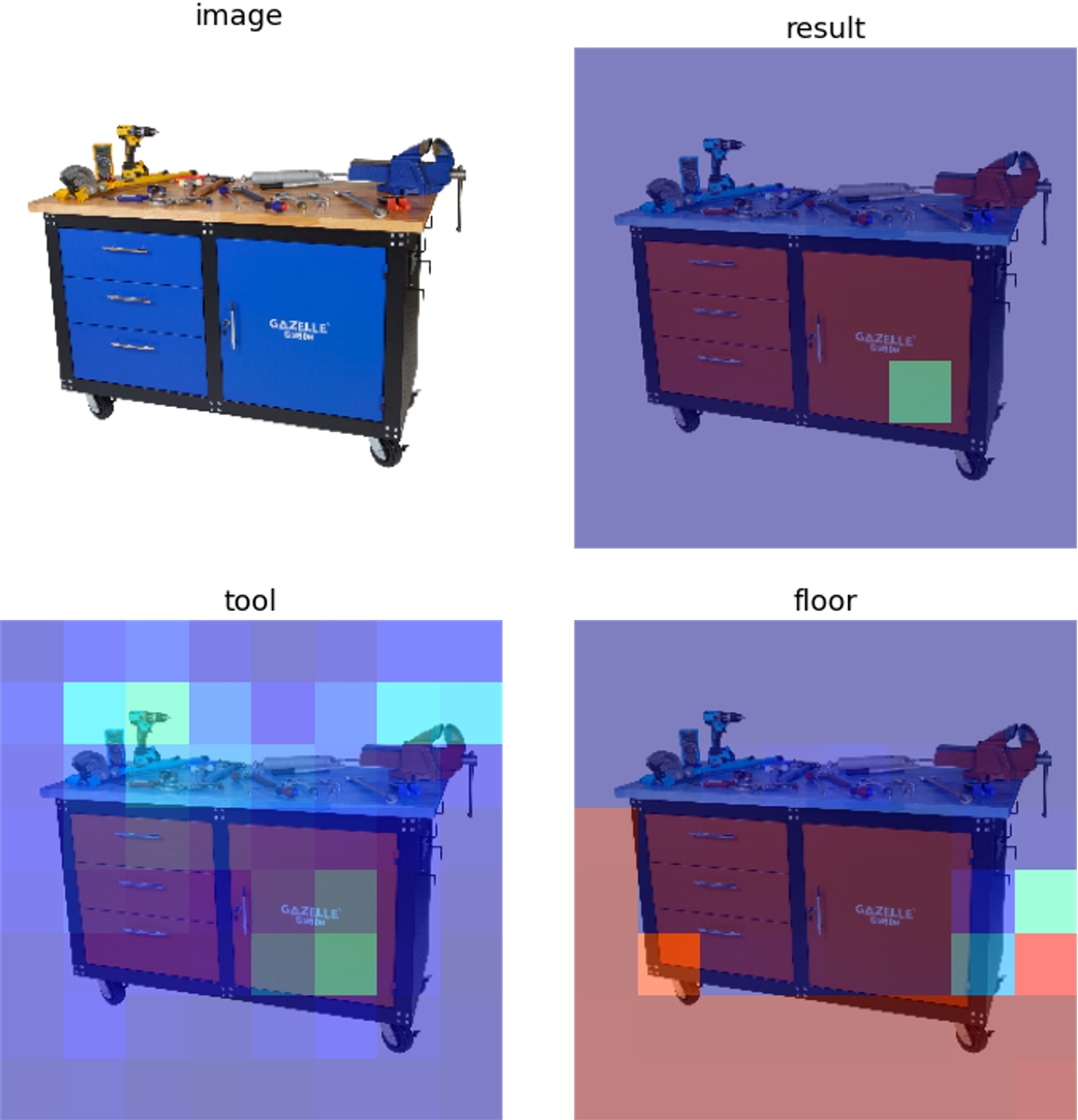}
         \caption{False positive}
         \label{fig:tool_fp_a}
     \end{subfigure}
     \hfill
     \centering
     \begin{subfigure}[b]{0.47\textwidth}
         \centering
         \includegraphics[width=\textwidth]{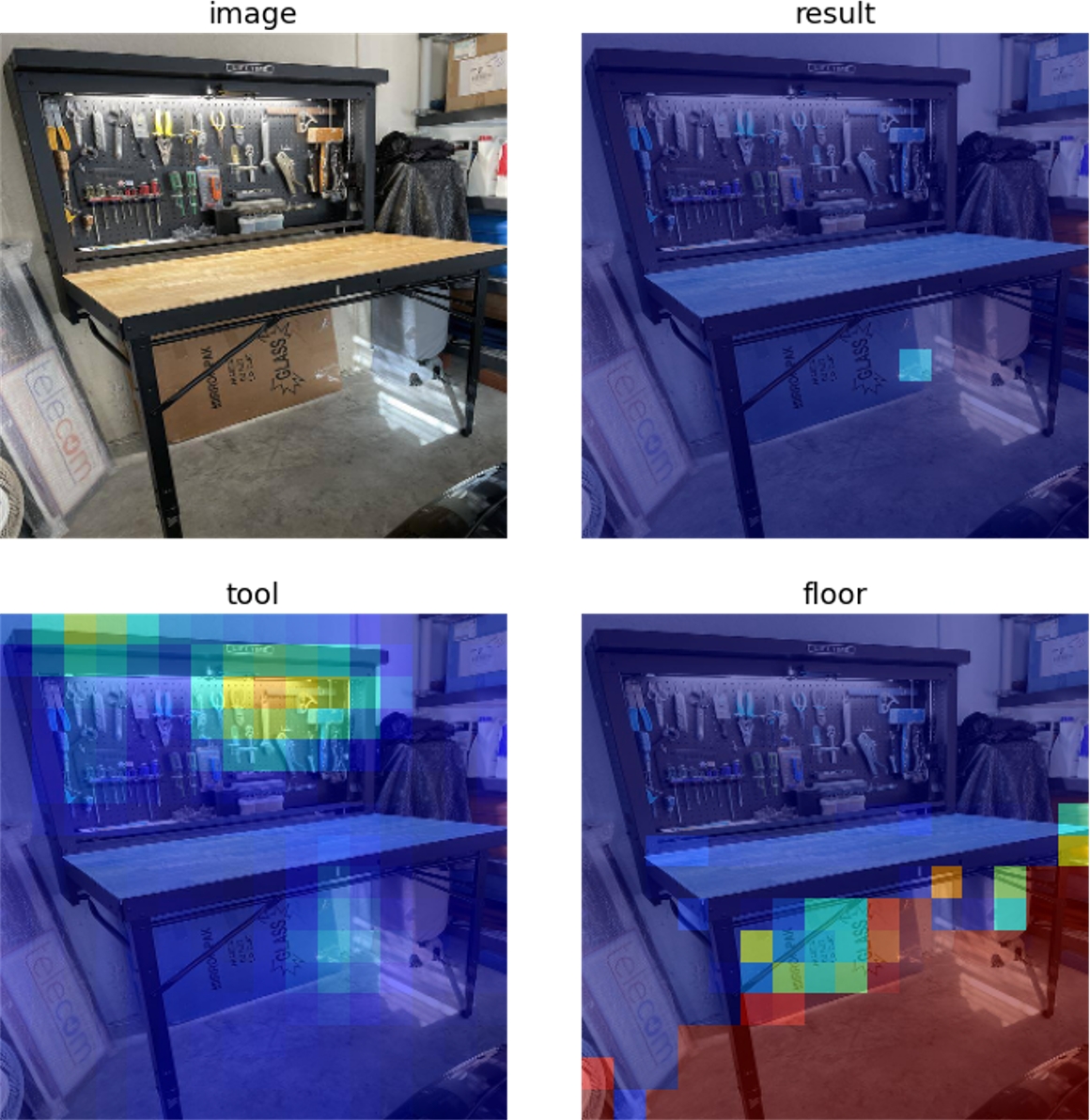}
         \caption{False positive}
         \label{fig:tool_fp_b}
     \end{subfigure}
     \\
     \centering
     \begin{subfigure}[b]{0.47\textwidth}
         \centering
         \includegraphics[width=\textwidth]{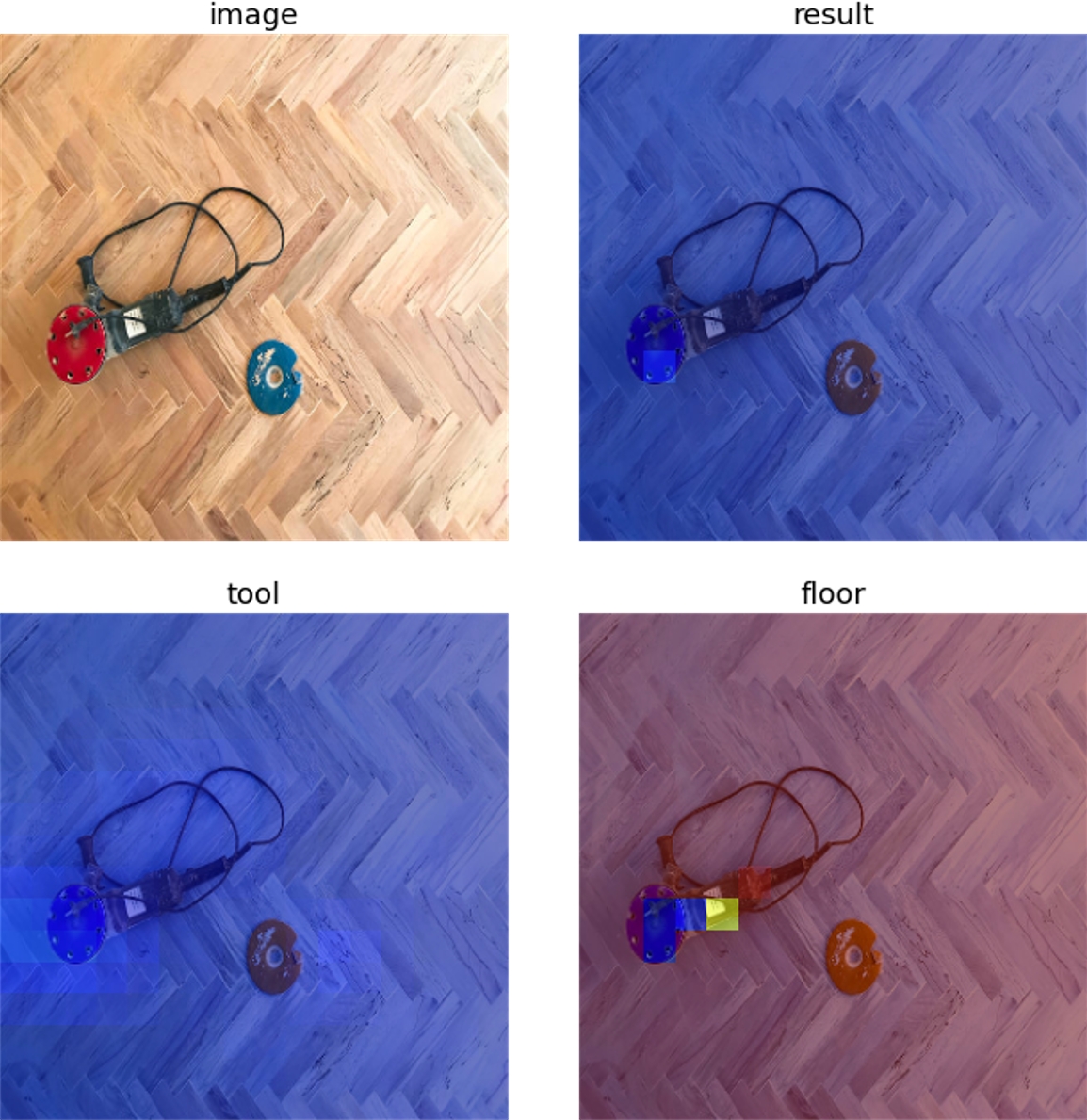}
         \caption{False negative}
         \label{fig:tool_fn_a}
     \end{subfigure}
     \hfill
     \centering
     \begin{subfigure}[b]{0.47\textwidth}
         \centering
         \includegraphics[width=\textwidth]{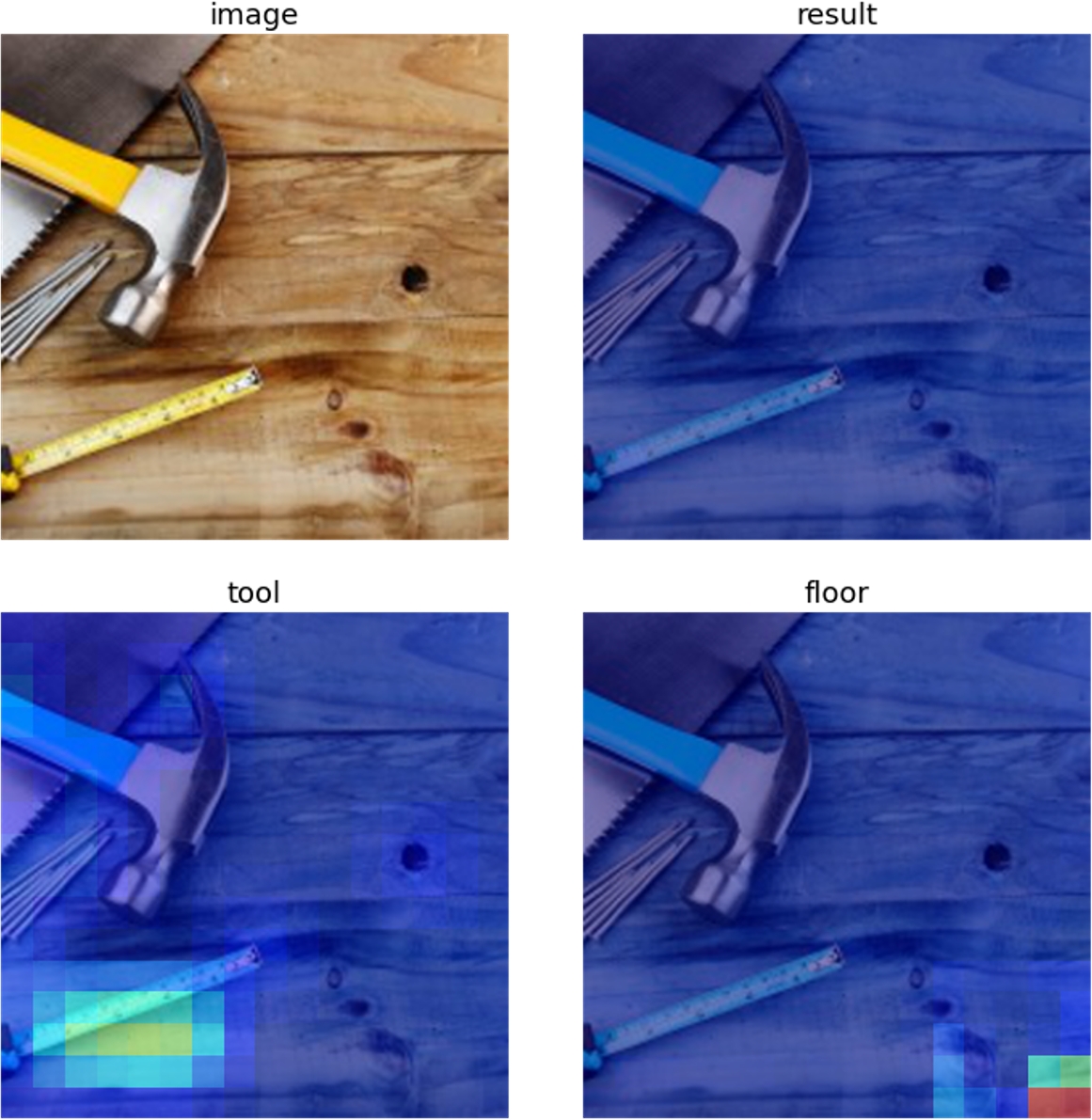}
         \caption{False negative}
         \label{fig:tool_fn_b}
     \end{subfigure}
     \caption{Tool on floor: errors.}
     \label{fig:tool_fp}
\end{figure}

The two false positives are shown in Figure \ref{fig:tool_fp_a} and \ref{fig:tool_fp_b}. The neuro-symbolic program incorrectly reasons that these cases are a tool left on the floor (false positives). This is due to errors in the symbols. There is a wrong association of the symbol tool in both images. On the left, the Gazelle logo is associated with a tool. We hypothesize that the reason for this flaw is that Gazelle is a manufacturer of bicycles: many images on the web about this brand involve tools. The language-vision model probably has learned a bias to associate Gazelle with tools. On the right, the duct tape is considered to be a tool. From a semantic point of view this makes sense. These symbol errors propagate into the reasoner's outputs. Refining the prompts that we pose to the language-vision models, may overcome such errors in the symbols.

Figure \ref{fig:tool_fn_a} and \ref{fig:tool_fn_b} show missed cases (false negatives). Again, the source of the errors is in the symbols. On the left, the tool (a grinder) is not recognized as such. This is a flaw in the language-vision model, probably because this tool does not appear often in everyday images and language. On the right, the floor is not recognized as such. We hypothesize that this is due to a lack of context within the image: it could also be a wooden plate. Without the proper evidence for each involved symbol, the reasoner cannot assess these configurations correctly. The computation time per image is approximately 1 second on a standard CPU without any optimization of efficiency.

\subsection{Leaking Pipe}

The second experiment is about another task: to find a leaking pipe. We collected 15 images of very different cases with various pipes and leaking fluids. Likewise the abandoned tool, we include various viewpoints and distances to the scenes. 

\begin{figure}[H]
    \centering
     \begin{subfigure}[b]{0.47\textwidth}
         \centering
         \includegraphics[width=\textwidth]{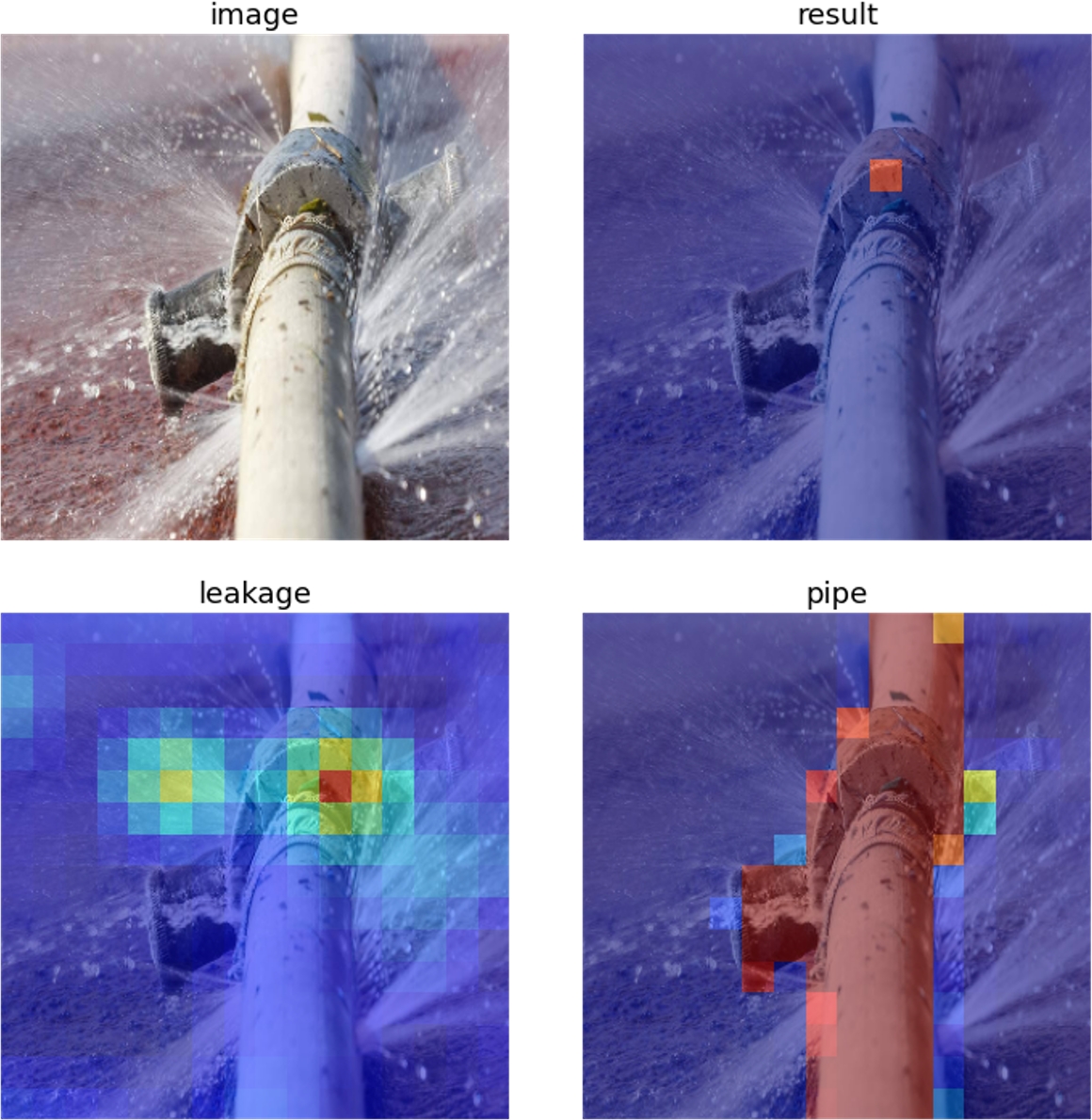}
         \caption{True positive}
         \label{fig:pipe_tp_a}
     \end{subfigure}
     \hfill
     \begin{subfigure}[b]{0.47\textwidth}
         \centering
         \includegraphics[width=\textwidth]{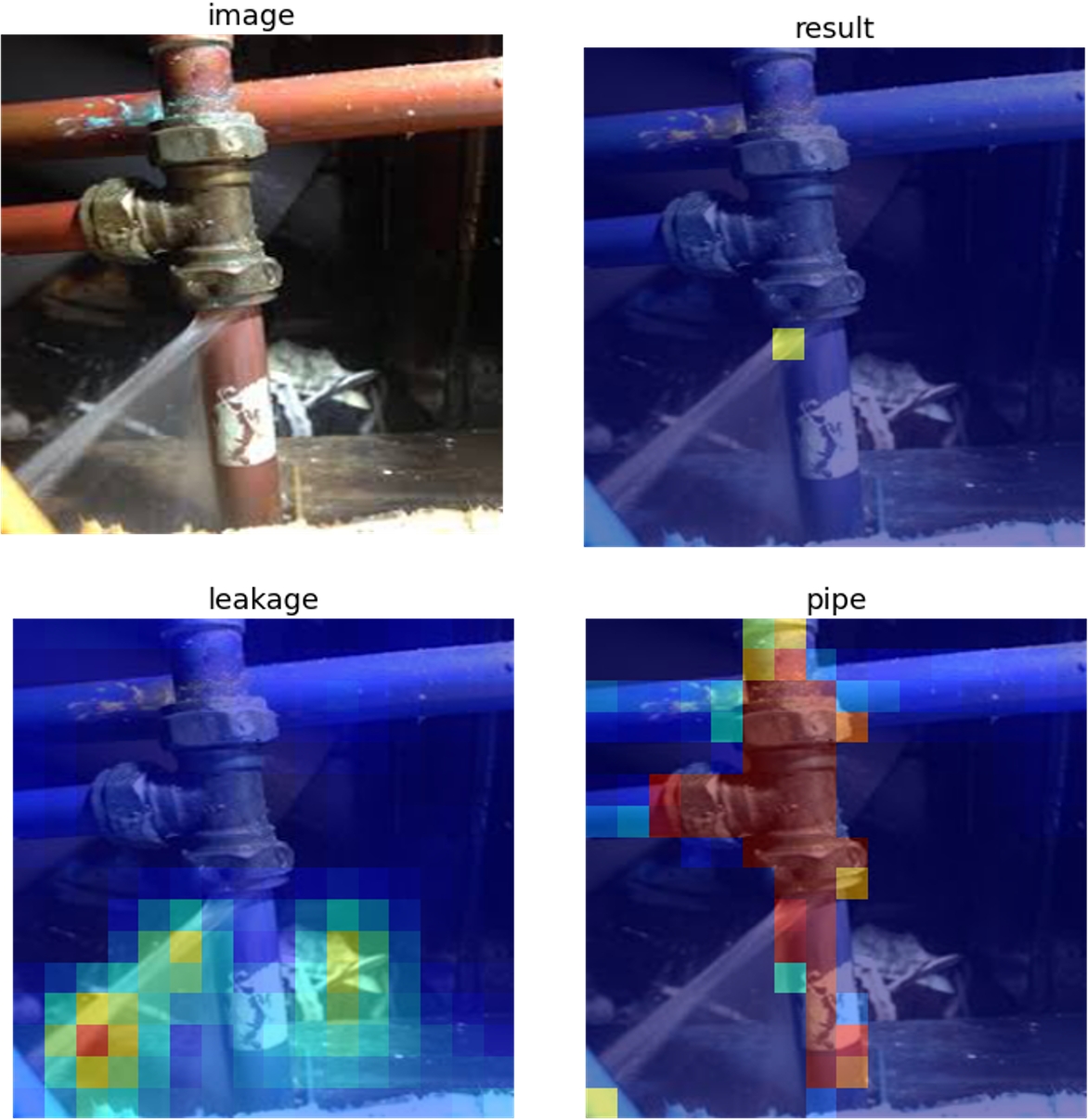}
         \caption{True positive}
         \label{fig:pipe_tp_b}
     \end{subfigure}
     \centering
     \begin{subfigure}[b]{0.47\textwidth}
         \centering
         \includegraphics[width=\textwidth]{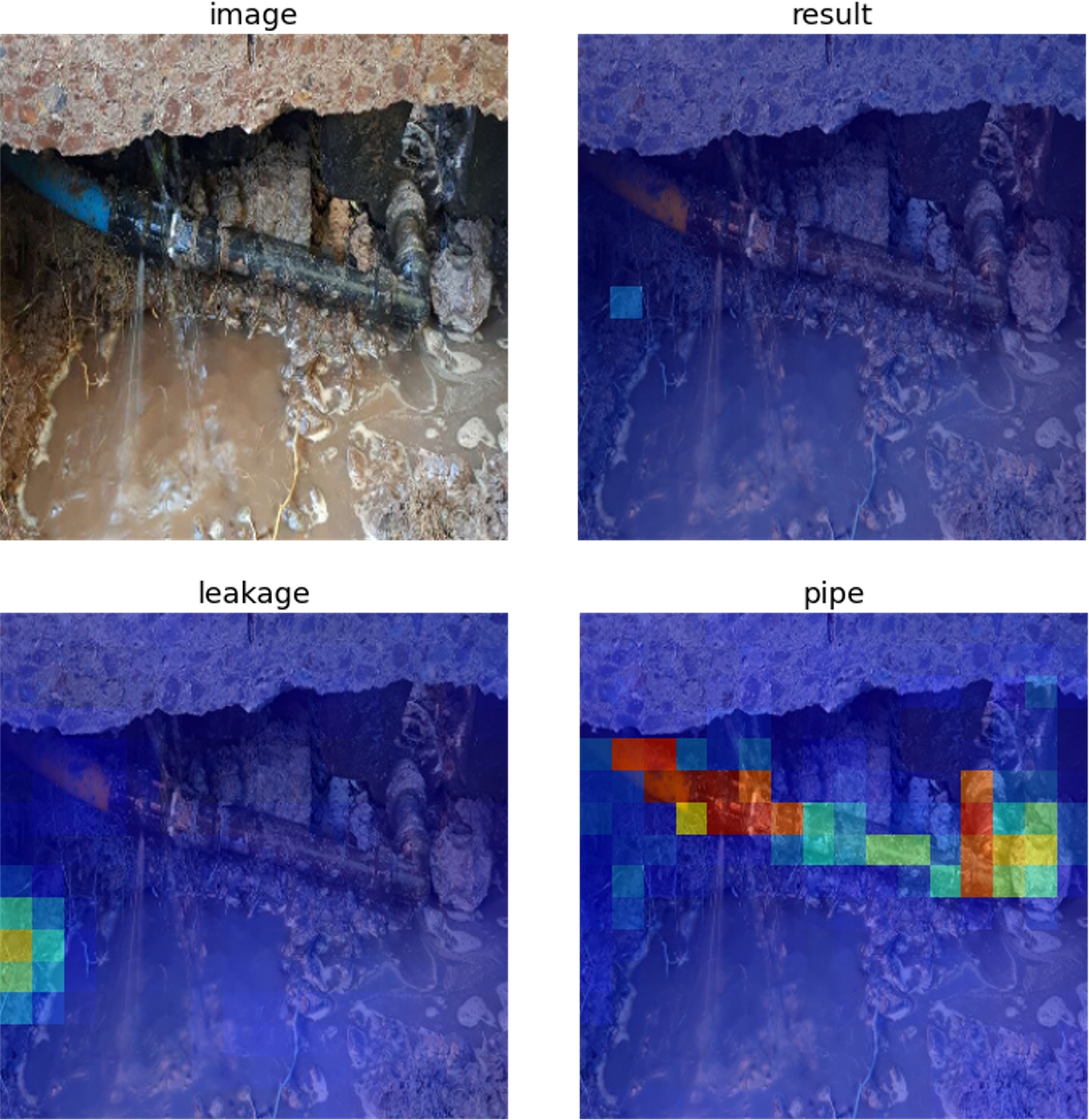}
         \caption{False negative}
         \label{fig:pipe_fn_a}
     \end{subfigure}
     \hfill
     \centering
     \begin{subfigure}[b]{0.47\textwidth}
         \centering
         \includegraphics[width=\textwidth]{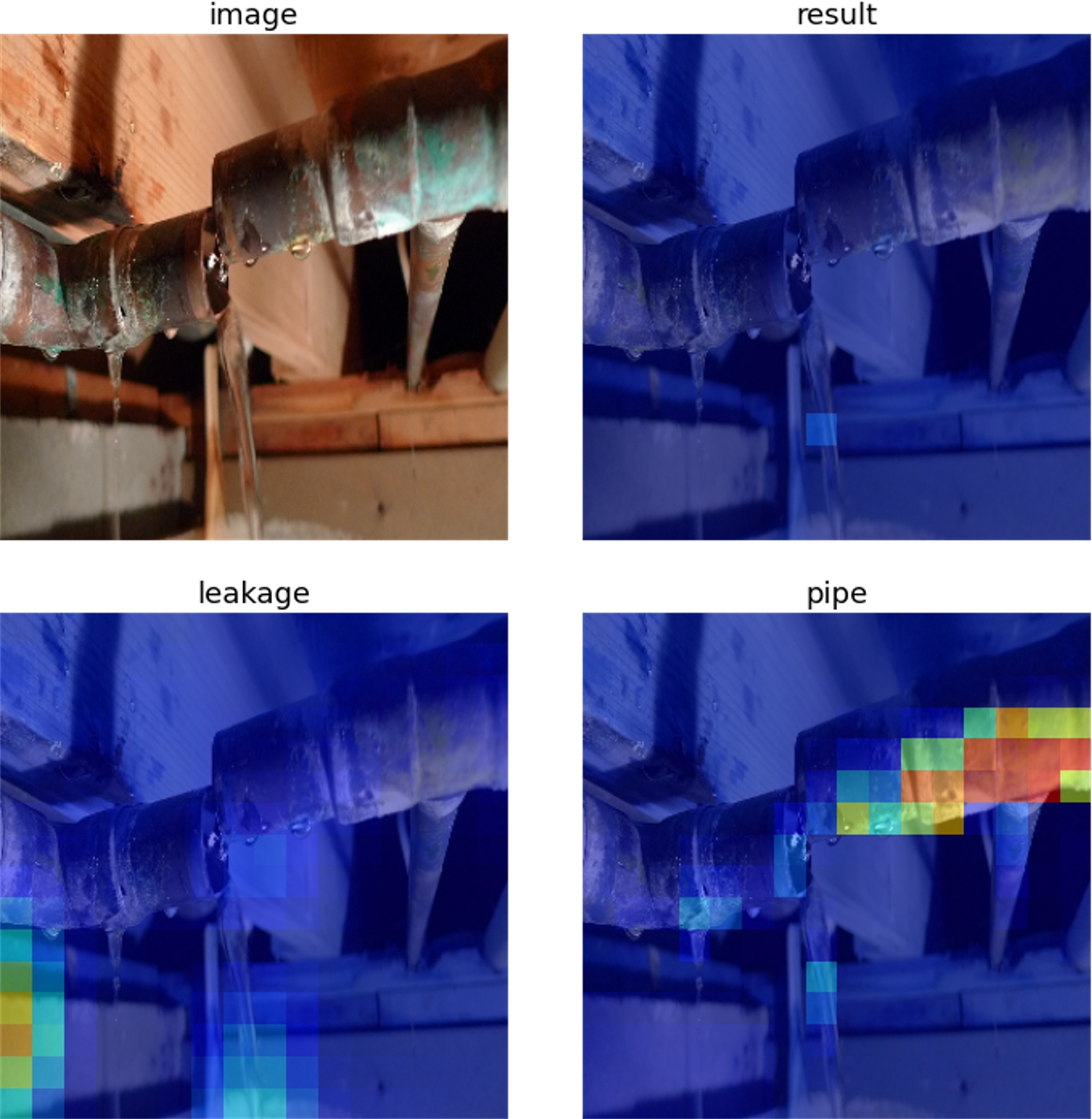}
         \caption{False negative}
         \label{fig:pipe_fn_b}
     \end{subfigure}
     \caption{Leaking pipe.}
     \label{fig:pipe_res}
\end{figure}

There are 15 positives, from which we detect 13 cases. Figure \ref{fig:pipe_tp_a} and \ref{fig:pipe_tp_b} show 2 out of those 13 cases. The organization of the figure is the same as previous result figures. Although the scenes are very different, the symbols are predicted well. Often the pipes and leakages have a high probability at the respective symbols. The neuro-symbolic program is able to pinpoint a place in the image where the spatial configuration is fulfilled (bright spot in the heatmap). 

Out of the 15 actual cases, 2 were considered to be a negative (miss). Figure \ref{fig:pipe_fn_a} shows a false negative where the leakage was localized too far away from the pipe. Hence the spatial arrangement did not fulfil the definition and therefore it was not assessed as a leaking pipe. In \ref{fig:pipe_fn_b} there is a false negative, because the left part of the broken pipe was not assessed as such. Therefore the leakage is not close to a part in the image that was considered as a pipe.

\subsection{Performance and Ablations} 

\begin{figure}[H]
    \centering
    \includegraphics[width=\textwidth]{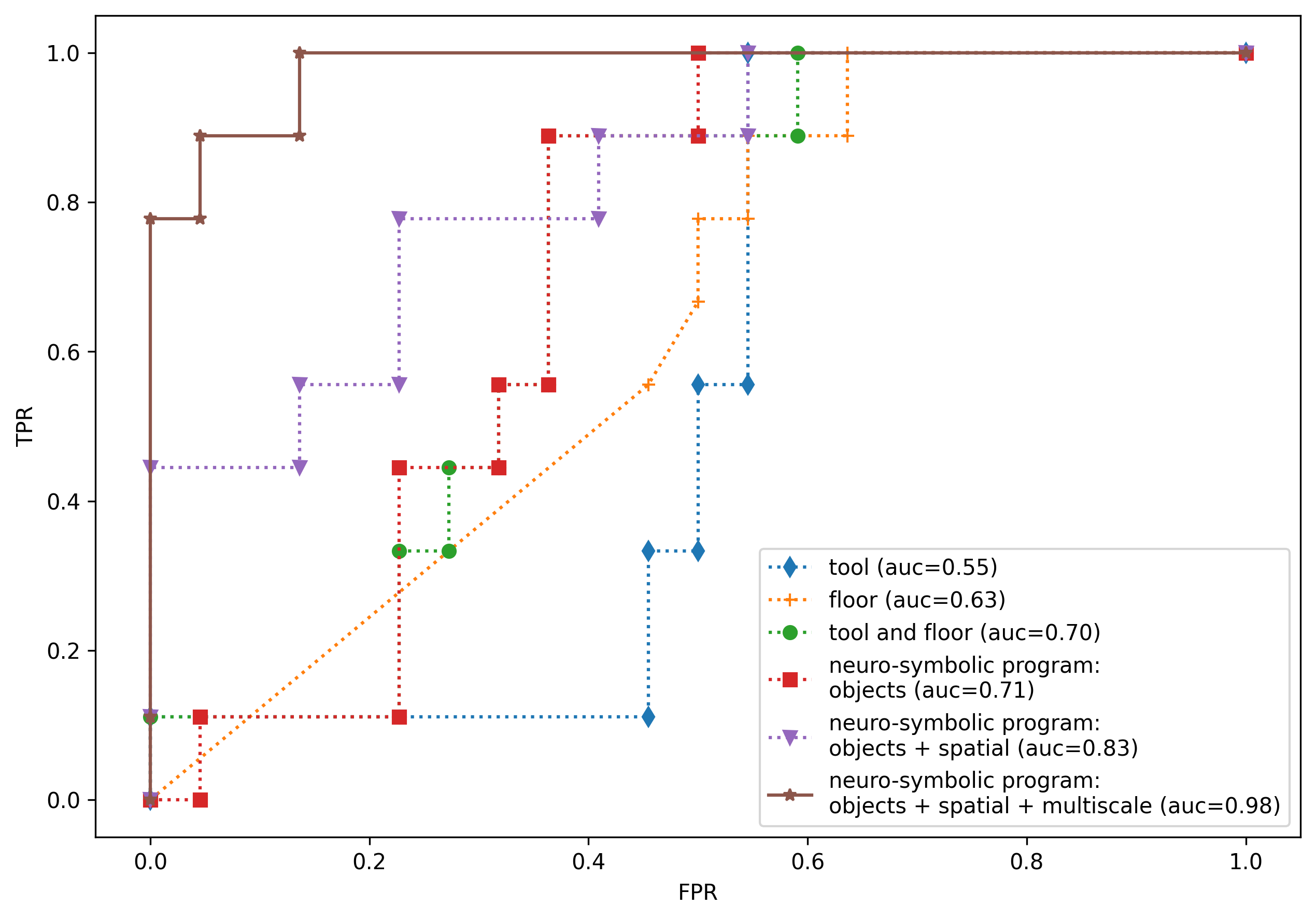}
    \caption{ROC curves for abandoned tool on floor. The neuro-symbolic program is more effective than alternative combinations of tool and floor. Spatial information and multi-scale reasoning are helpful.}
     \label{fig:rocs}
\end{figure}

The performance is evaluated by ROC curves as shown in Figure \ref{fig:rocs}. Our neuro-symbolic program is compared against baselines that use partial information (only the tool or only the floor) and a baseline that uses the same information (tool and floor) but without the spatial configuration. For ablation, three variants of the neuro-symbolic program are evaluated, each having more contextual knowledge and multi-scale reasoning. 

It can be concluded that both tool and floor are required, and the spatial configuration is also essential. With only tool and floor as inputs, the performance becomes almost random, resp. AUC=0.55 and AUC=0.63. With both inputs, yet without spatial configuration, the performance increases only slightly: AUC=0.70. Including tool and floor in the neuro-symbolic program, without taking their spatial relations into account, is equally ineffective: AUC=0.71. When the spatial configuration is considered by the neuro-symbolic program, the performance increases significantly: AUC=0.83. Including multi-scale makes the neuro-symbolic program very effective: AUC=0.98. Most of the situations of interest can be detected without false positives, whereas the alternatives produce many false positives.



\section{Conclusions}

For open world settings, we proposed a method to find spatial configurations of multiple objects in images. It enables expert-driven localization of new or unseen object configurations. Our method is able to find situations of interest in a robotic inspection setting: abandoned tools on floors and leaking pipes. The tools, floors, pipes and leakages have not seen before and no task-specific training was performed. Most of the situations of interest were correctly localized. A few false positives occurred, due to erroneous object proposals. This was caused by a bias in the language-vision model, e.g., a logo of a tool that was considered to be a tool. A few false negatives happened, due to missed object proposals. A typical example is a close-up image of a floor, which was missed because context was lacking. Our method avoids the necessity of learning dedicated models for each of the involved objects, which makes our method flexible and quickly operational in an open world.

%
%
%
\bibliographystyle{splncs04}
\bibliography{refs}

\end{document}